%% file: root.tex
\documentclass[a4paper, 10pt, conference]{ieeeconf}      % Use this line for a4 paper

\IEEEoverridecommandlockouts                              % This command is only needed if 
\usepackage{cite}
\usepackage{graphicx}
\usepackage{caption}
\usepackage{amsmath} % assumes amsmath package installed
\usepackage{amssymb}  % assumes amsmath package installed
\usepackage{url}
\usepackage{xcolor}
\usepackage{algorithm,algorithmic}
\usepackage{subfig}

\title{\LARGE \bf Accurate Global Trajectory Alignment using Poles and Road Markings}

\author{
Haohao Hu$^1$, Marc Sons$^2$ and Christoph Stiller$^1$% <-this % stops a space
\thanks{$^1$Author is with Institute of Measurement and Control Systems, Karlsruhe Institute of Technology, Karlsruhe, Germany. {\tt\small haohao.hu@kit.edu}}%
\thanks{$^2$Author is with FZI Research Center for Information Technology, Karlsruhe, Germany. {\tt\small sons@fzi.de}}%
}

\begin{document}
\maketitle
\thispagestyle{empty}
\pagestyle{empty}

%% Main contents
\input{abstract}
\input{introduction}
\input{related_work}
\input{algorithm_overview}
\input{evaluation}
\input{conclusions}
\input{acknowledgment}

%% Bib refs
\bibliographystyle{IEEEtran}
\bibliography{IEEEabrv,references}
\end{document}

%% file: abstract.tex
\begin{abstract}
Currently, digital maps are indispensable for automated driving.
However, due to the low precision and reliability of GNSS particularly in urban areas, 
fusing trajectories of independent recording sessions and different regions is a challenging task.
To bypass the flaws from direct incorporation of GNSS measurements for geo-referencing, 
the usage of aerial imagery seems promising. 
%
% However, this can be however easily solved by an excellent geo-referencing. 
%
Furthermore, more accurate geo-referencing improves the global map accuracy and allows to estimate the sensor calibration error.\\
%
% Using geo-referencing, quality of digital maps can be improved obviously.
%
In this paper, we present a novel geo-referencing approach to align trajectories to aerial imagery using poles and road markings.
To match extracted features from sensor observations to aerial imagery landmarks robustly, a RANSAC-based matching approach is applied in a sliding window.
For that, we assume that the trajectories are roughly referenced to the imagery which can be achieved by rough GNSS measurements from a low-cost GNSS receiver.
Finally, we align the initial trajectories precisely to the aerial imagery by minimizing a geometric cost function comprising all determined matches.    
Evaluations performed on data recorded in Karlsruhe, Germany show that our algorithm yields trajectories which are accurately referenced to the used aerial imagery.
\end{abstract}

%% file: introduction.tex
\section{INTRODUCTION}
\label{sec:INTRODUCTION}
Using digital maps is substantial for automated driving at the present time.
%
% Even though research work has been done for perception module of automated vehicle, the perception is still not accurate and reliable enough. 
%
To achieve reliable results from the perception, behaviour generation or planning module in automated vehicles,
processed information from current sensor readings is merged and validated with information from digital maps.
Actually, maps without any global reference are sufficient to solve the aforementioned tasks since it is not necessary to know exactly where the ego vehicle is in the world but 
relatively to the map \cite{sons2017mapping}. 
However, maps generated without exact and absolute global reference measurements usually provide a high local accuracy but worse global consistency.
Such maps often show drift and scale errors which reduce the map quality especially when multi-modal and model based filter algorithms are applied for localization.
Geo-referencing can be used to correct those errors and improve the global consistency.
Furthermore, city-scaled areas are usually mapped from multiple independent recording sessions \cite{Sons2018MultiDriveMapping}.
To achieve a robust and reliable fusion of all recording sessions into one global frame, an accurate global reference is fundamental. 
The easiest and most intuitive way to reference driven trajectories globally is to incorporate GNSS measurements into the mapping process.
\begin{figure}[thpb]
\centering
\includegraphics[width = \columnwidth]{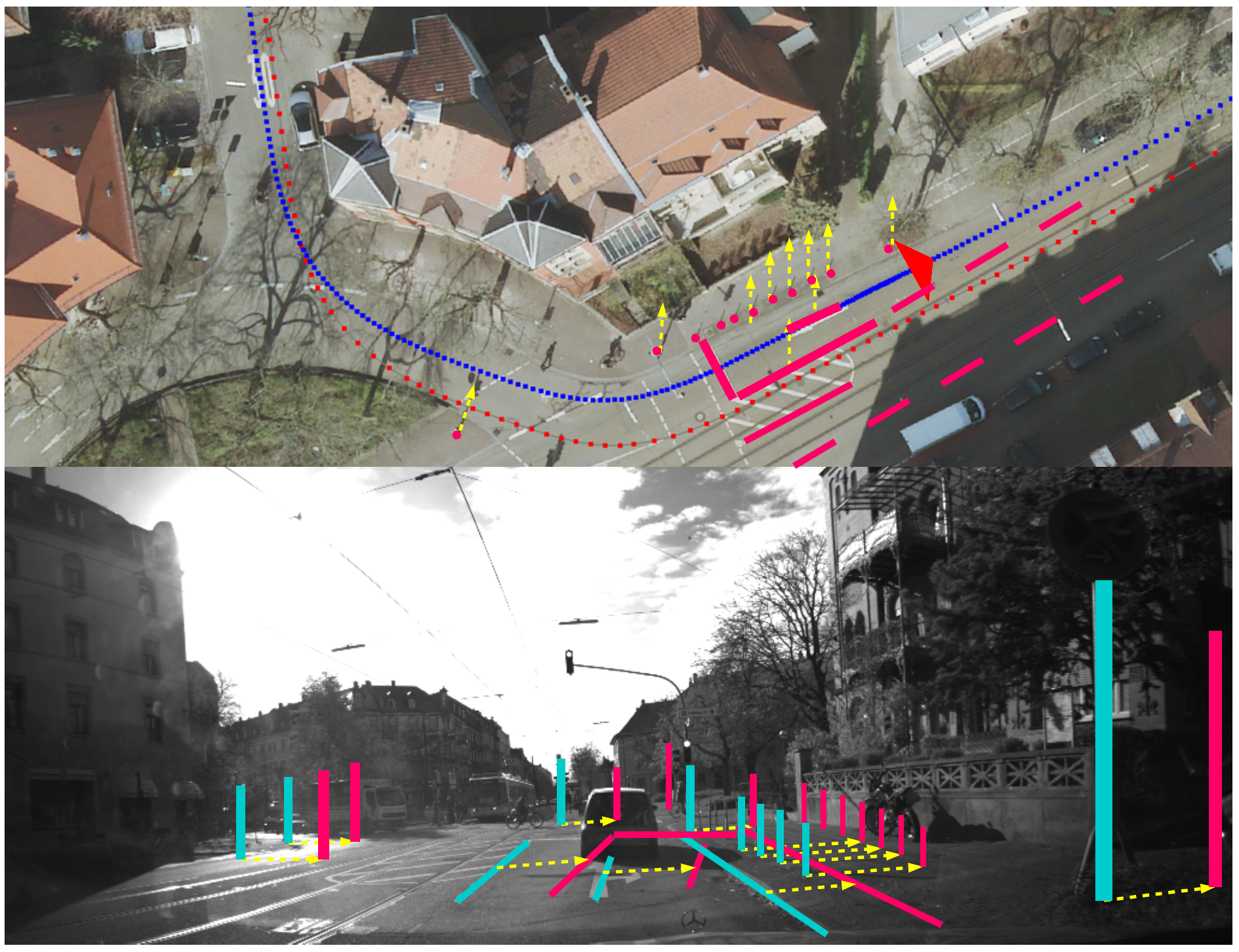}
\caption{
Depiction of the idea and results of our approach. 
The top image shows the initial (red) and our post-aligned trajectory (blue). 
Obviously, the alignment fits well to the road geometry of the underlying aerial imagery.
The red triangle shows the view point of the bottom image which shows the displacement (yellow arrows) of pole and road marking features (cyan) which we matched 
between vehicle sensor readings and aerial imagery landmarks (pink) to align the trajectories accurately. 
Aerial Imagery: \copyright Stadt Karlsruhe | Liegenschaftsamt
}
\label{fig:motivation}
\end{figure}
However, due to multipath-, shadowing- and atmospherical drift issues especially in urban areas,
only a rough and inconsistent global map accuracy and -reference can be achieved.
%
% In this case, geo-referencing can provide for each piece of map a more precise and reliable global reference.\\
%

%
In this work, we present a novel approach to align trajectories from multiple recording sessions precisely to aerial imagery using pole-like infrastructures and road markings.
As input, we assume jointly estimated and roughly geo-referenced trajectories \cite{Sons2018MultiDriveMapping}.
Since current automated vehicles are usually equipped with cameras und spinning laserscanners, we, furthermore, assume the availability of frequently point cloud and image 
measurements from each recording session.
To improve the global accuracy and geo-referencing, we align the initial trajectories precisely to geo-referenced aerial imagery using feature matches 
between the vehicle sensor- and the aerial imagery domain.
%
% On the one hand, we extract poles and road markings from vehicle sensor readings.
% %
% On the other hand, we extract the same type of features from aerial imagery.
% %
% Pole-like structures as well as road markings are clearly visible and, hence, easily detectable in both domains.
% %
% 
% 
% 
% 
% 
% 
% %
Pole-like structures can be robustly extracted from point clouds and, additionally, distinguished between trees and traffic signs 
by semantic labelling in the images and successive label transfer into the point cloud domain.
Furthermore, road markings can be precisely detected and classified in the images.
Since these features are also clearly visible in aerial imagery, provide a good coverage in urban areas and are persistent over long time-periods,
we choose them for our approach.
To match features robustly between both domains, we propose a novel sliding window approach.
We determine matches in each local window robustly by evaluating randomly selected match hypotheses in a RANSAC scheme \cite{fischler1981random} using different distance measures.
Finally, all matches are treated as constraints in a non-linear least squares (NLS) adjustment problem 
whose minimization yields the precise global alignment of the initial trajectories.
In summary, we propose a method to postprocess trajectories provided from any multi- or single-session outdoor mapping approach to 
achieve accurately geo-referenced and globally correct trajectories. 
By that, we bypass the need of reliable and accurate GNSS measurements. 
% In summary, we propose a solution to align trajectories provided from an outdoor SLAM or odometry approach accurately to geo-referenced aerial imagery 
% and by that, bypass the need of reliable and accurate GNSS measurements. 
%
Our main contributions are:
\begin{itemize}
\item a selection of geometric features which are on the one hand good observable in the sensor- and aerial imagery domain and on the other hand, 
establish a good coverage in typical inner-city and sub-urban scenarios.
\item a novel RANSAC-based sliding window algorithm to match features robustly between both domains.
%
% \item a
\end{itemize}
%
% This paper is structured as follows: Firstly, we give an overview on the work related on this topic in Section \ref{sec:RELATED WORK}.
% %
% In section \ref{sec:ALGORITHM OVERVIEW}, we present our trajectory geo-referencing algorithm. 
% %
% Afterwards, the experiments and achieved results are evaluated and discussed in Sec. \ref{sec:Evaluation}.
% %
% Finally, we conclude our work and give an outlook in Section \ref{sec:Conclusion_and_future_work}.  
%

%% file: related_work.tex
\section{RELATED WORK}
\label{sec:RELATED WORK}
This section reviews state of the art approaches related to this work.
Leung et al. \cite{Leung2008GroudImageWithAerialImageLoc} presents a monocular vision based particle filter localization in urban environments using aerial imagery as a reference map.
Image processing techniques like Canny edge detection and a progressive probabilistic Hough transform are used to create a line feature map from aerial imagery.
For localization, line features are detected from ground-based monocular camera image and used as observations.
The approach achieves an positioning accuracy of several meters which is a magnitude worse as the accuracy achieved with our approach.
Ding et al. \cite{Ding2008LidarWithAerialImage} analyzes vanishing points to detect 2D corner features from aerial imagery.
The same kind of features are also extracted from LiDAR depth maps.
A Hough transform and a M-estimator are applied to obtain matches between both domains.
Finally, the camera pose is estimated using the obtained 2D corner correspondences.
Tournaire et al. \cite{Tournaire2006TowardsAS} proposes an image-based approach for ground-based image to aerial imagery geo-referencing, 
called "GCO-based geo-referencing (Ground Control Objects)".
A Canny edge detector and several basic image processing steps are performed to extract road markings from aerial imagery and ground-based images.
Afterwards, an "analysis-synthesis" based matching approach is used for alignment.
Instead of using road markings, Bansal et al. \cite{Bansal2012FacadeMatching} uses facades to match street-level images to a database of airborne images.
Busch et al. \cite{Busch2018HDMapLidar} generates a lane-level high definition map of a complex crossing from several static mounted 3D laser scanners.
Trajectories provided from dynamic objects passing this crossing are processed within a clustering scheme and a least squares adjustment is finally used to 
generate a road network automatically.
To get the absolute location of the generated map, they extract poles from pointclouds as observations and label them in an aerial image manually as landmarks.
An ICP based approach is applied to align the automatically detected poles to the manually measured poles.
The approach achieves a root mean square error of 0.05 m between detected and measured poles.     
K{\"u}mmerle et al. \cite{Kmmerle2011LargeSG} provides a Graph-SLAM approach utilizing publicly accessible aerial imagery as prior information 
to improve the global consistency of maps.
%
% This approach is able to achieve an average error of 0.85 m and the graph-based SLAM algorithm without prior information achieved an average error of 1.3 m.
%
% And the standard deviation of the estimated distances is substantially smaller than the algorithm without using prior information.
%

%
All mentioned approaches align ground-level images to geo-referenced aerial imagery to estimate an absolute ego sensor pose 
or to stitch all images together.
%
% Busch et al. \cite{Busch2018HDMapLidar} uses the poles to estimate the absolute location of a from lidar sensors automatically generated road network with aerial imagery.
%
% K{\"u}mmerle et al. \cite{Kmmerle2011LargeSG} uses aerial imagery as prior information of Graph-SLAM to improve the map global consistency.
%
To our best knowledge, there is no existing work which align whole trajectories from multiple independent recording sessions consistently using geo-referenced aerial imagery
and multi-modal sensor measurements.
%
% But the trajectory geo-referencing process can improve the global consistency and accuracy of automated driving maps.
% %
% With geo-referencing, the several maps fusion problem can be also easily solved.
%
% Hence, we will present, evaluate and discuss an novel trajectory geo-referencing approach in this paper, which can align a roughly geo-annotated trajectory to an geo-referenced aerial imagery using poles and road markings.
% %
% The main contribution of our work is that we use an sliding window and RANSAC based approach to get accurate and reasonable feature association results, which can provide reliable feature correspondences for pose graph optimization process.
% %
% The geo-referenced trajectory can improve the automated driving maps accuracy obviously.

%% file: algorithm_overview.tex
\section{ALGORITHM OVERVIEW}
\label{sec:ALGORITHM OVERVIEW}
\begin{figure}
\centering
\includegraphics[width = \columnwidth]{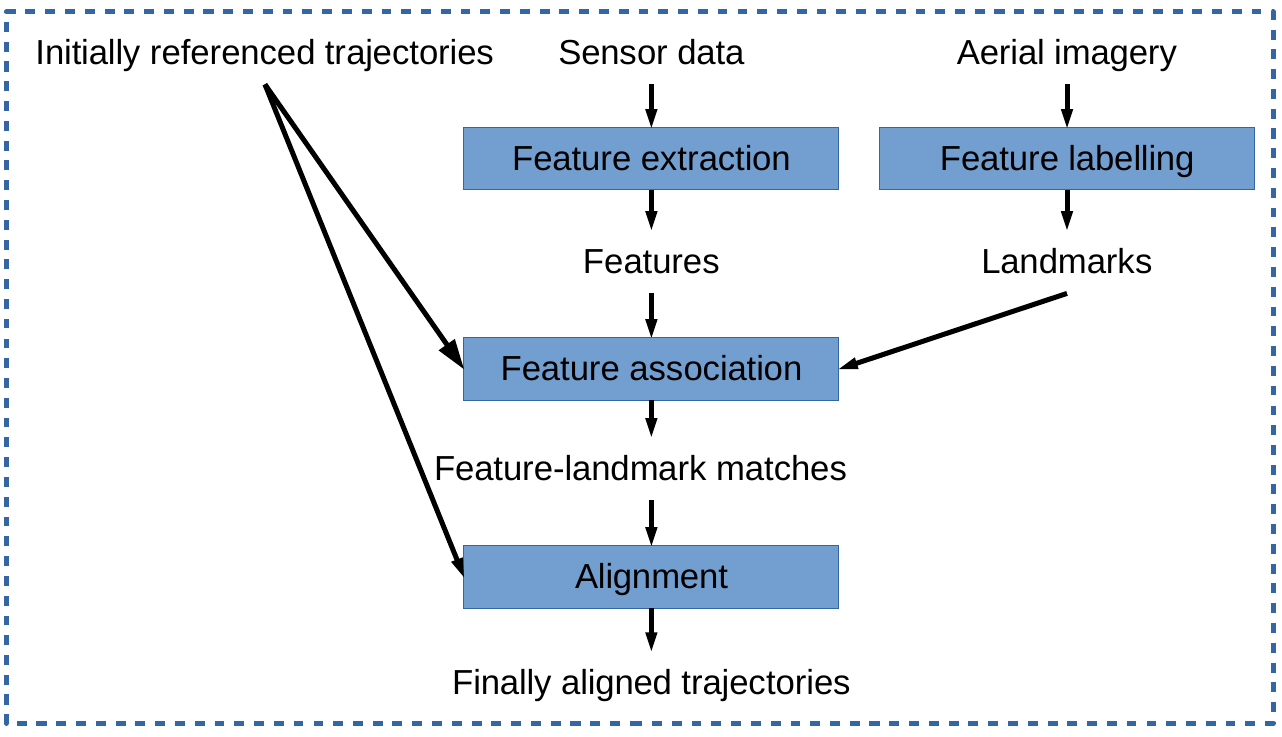}
\caption{Overview of the processing pipeline.}
\label{fig:approach}
\end{figure}
In this section, we present the details of our approach which comprises three main processing steps (see Fig. \ref{fig:approach}).
In the first step, features are automatically extracted from the recorded point clouds and images.
Additionally, corresponding features (called landmarks in the following) are manually labelled in aerial imagery.
Details of this step are described in Sec. \ref{subsec:Feature Extraction}.
After feature extraction, we robustly match features to landmarks within local areas using our RANSAC-based approach.
Sec. \ref{subsec:Feature Association} describes this matching step in detail.
Finally, we solve a NLS problem to achieve the final trajectories.  
This step is presented in Sec. \ref{subsec:Pose Graph Optimization}
\subsection{Feature Extraction}
\label{subsec:Feature Extraction}
\begin{figure}[thbp]
\centering
\includegraphics[width = \columnwidth]{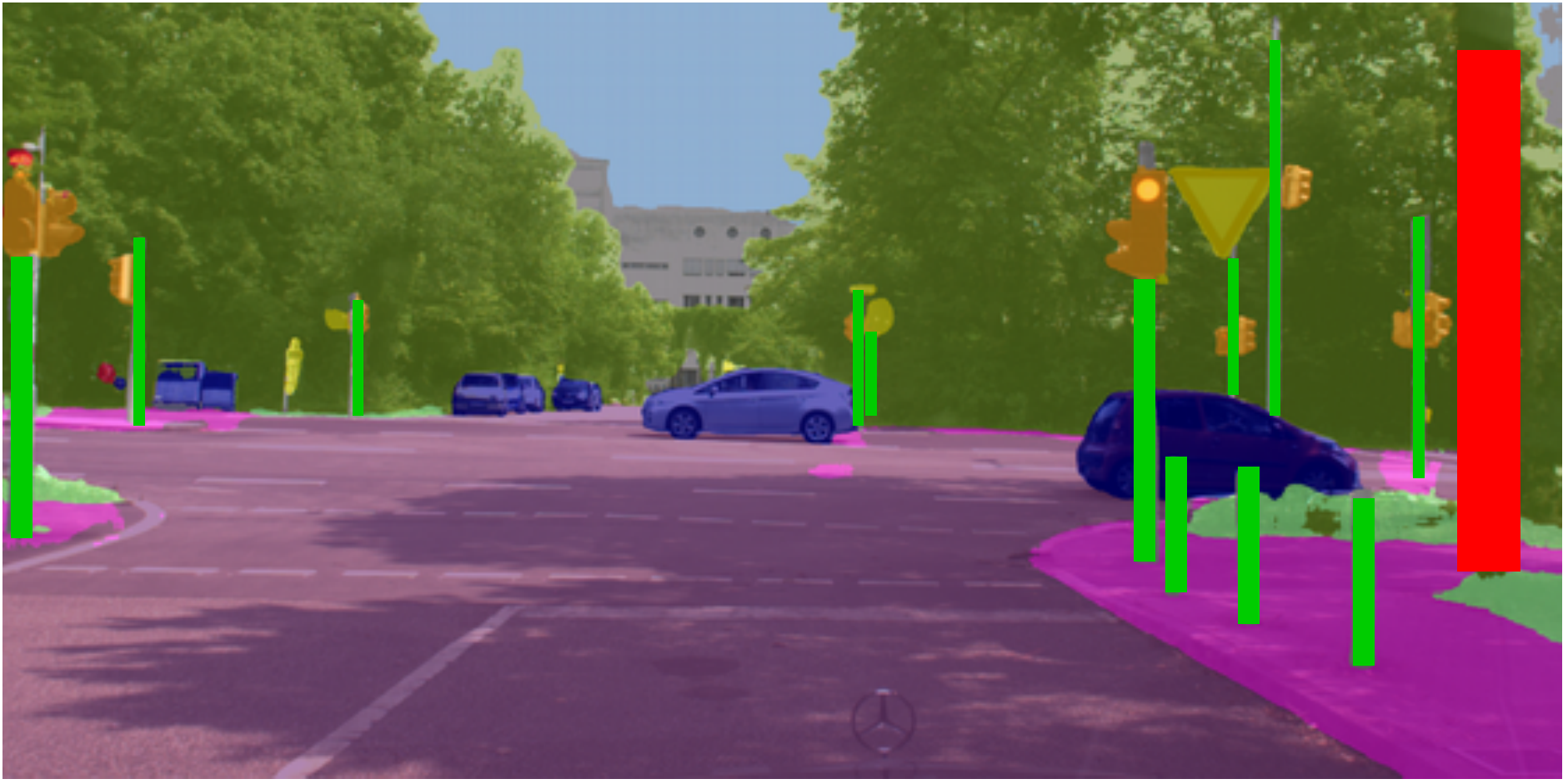}
\caption{
Exemplary depiction of the label transfer and the max voting wether a pole-like structure is used as feature or not. 
The green colored bars are accepted as feature since they were labeled as road signs, traffic lights or boundary posts whereas the red colored bar 
is classified as tree and rejected.}  
\label{fig:poles}
\end{figure}

This section describes the extraction and classification process of the used features.
We first describe the extraction and classification of pole-like infrastructures.
Thereafter, we discuss the recognition of road markings and -boundaries from the recorded images.   
For the extraction of pole-like infrastructures from the recorded point clouds, we use a similar approach as in \cite{sefati2017improving}.
% Firstly, an lidar point cloud based approach is applied to extract pole-like infrastructures.
%
This approach detects anything which appears like a pole in the point clouds. 
However, in the aerial imagery only road signs, advertising pillars, traffic lights and boundary posts are discernable.    
%
% Because of occlusion and chaos of trees and bad explicitness of building edges, the manual labelling of such features in aerial imagery is very hard.
%
% In contrast to trees and building edges, poles in aerial imagery can be easily labelled. 
%
Hence, we classify all detections and keep only the aforementioned types of pole-like infrastructures. 
For that, we classify the segmented points by label transfer from the image into the point cloud domain.
We utilize a modified ResNet38, which is trained using cityscapes-dataset \cite{Cordts2016Cityscapes} to obtain pixel-wise semantic labels from the recorded images. 
Each point of a pole segment is projected into the image using a known sensor calibration and associated with the corresponding pixel label.   
By max voting over all labels in a segment we obtain the pole label (see Fig. \ref{fig:poles}).
To detect and classify road markings, we apply the detection and classification approach proposed in \cite{Poggenhans2015RoadMarkings}.
Here, the detected road markings are classified as one of the following classes: arrows, stop-lines, pedestrian crossings, dashed and straight lines with different linewidth.
Additionally, boundaries between road and no road are extracted utilizing the semantic labels.
These boundaries improve the lateral alignment in areas with less road markings.
\subsection{Feature Matching}
\label{subsec:Feature Association}
\begin{figure}
\centering
\includegraphics[width = \columnwidth]{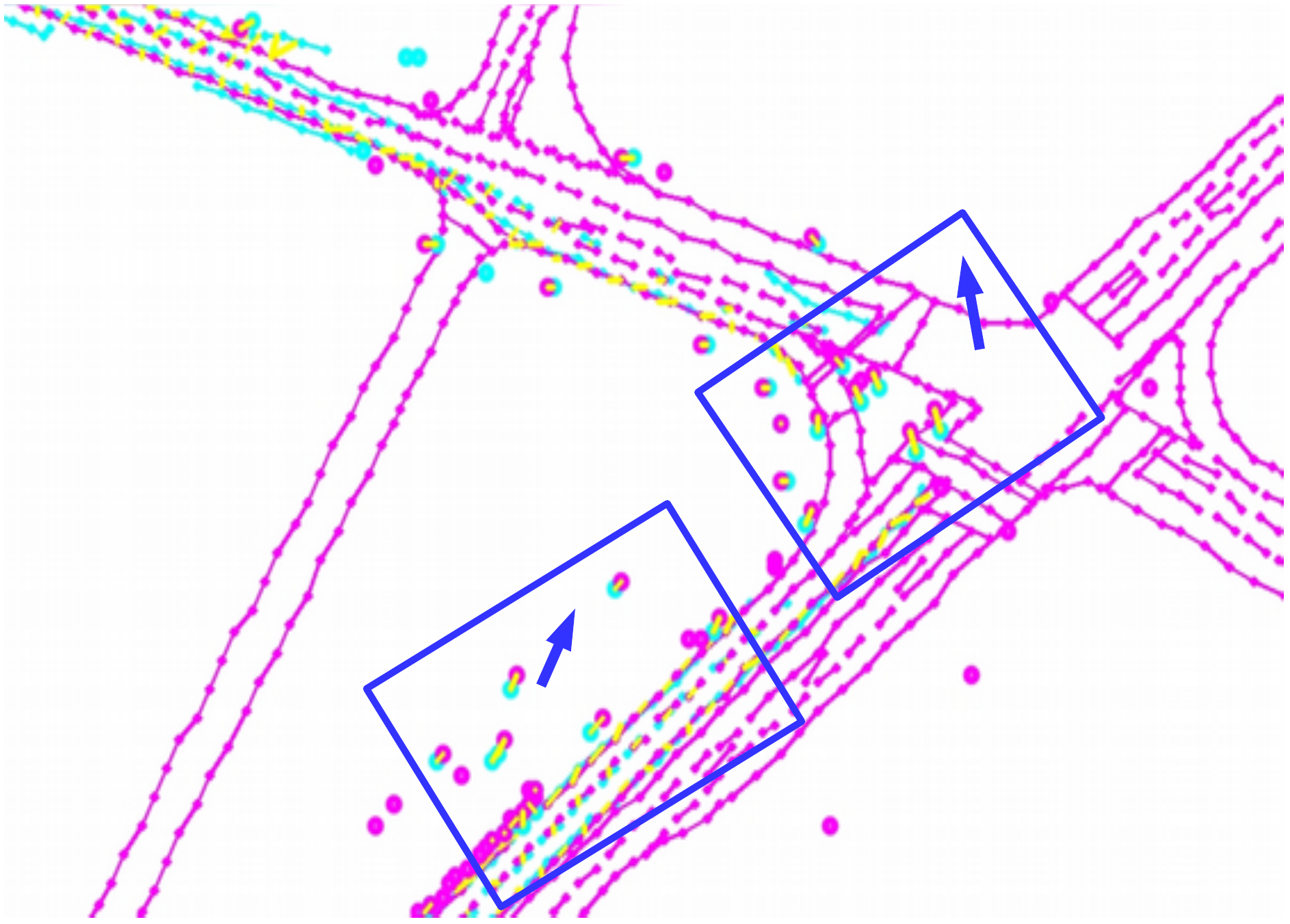}
\caption{Schematical depiction of pole and road marking landmarks (pink), pole features (cyan circles), 
road marking features (cyan line-segments), exemplary matching windows (blue boxes), determined matches (yellow lines)
and the computed displacement direction of the windows (blue arrows).}
\label{fig:association}
\end{figure}

As previously mentioned, the extracted features are clearly perceptible in all domains, however, 
except of rarely occuring arrows, stop-lines and pedestrian crossings, they lack of unique characteristics to match them robustly 
even with a reasonable prior referencing from GNSS. 
%
% To find feature correspondences between observations and landmarks, an feature association process is in this section presented.
%
% Our highly geo-referenced aerial imagery doesn't have hight information and to get the hight information of landmark labels, an another hight reference profile is used.   
%
Fig. \ref{fig:association} shows exemplary aerial imagery landmarks (pink) and, overlayed by the initial trajectories, detected 
poles (cyan circles) and road markings (cyan line-segments) at a typical crossing.
Furthermore, the yellow lines show the displacement vectors of correct matches. 
Obviously, the feature-landmark displacement is similar within local regions (blue boxes) and varies for different local regions.
%
% From the yellow lines, which connect feature correspondences, is clear, that the shifts from observation to landmark are different 
%
Therefore, we determine correct matches by estimating a transformation which minimizes the local displacement within a window $W$.
Thereby, the window $W$ is shifted along all mapped areas while keeping an overlap between neighbouring windows.
For matching, poles are represented by their intersection points with the local ground surface.
The distance measure for association and optimization is the euclidean distance $d_e(a)$ in this case.
Here, $a$ denotes an match between a detected pole feature and a pole landmark.
Furthermore, road markings and -boundaries are presented as line segments. % and road markings like  are sampled with a pre-setted length as line segments.

As distance measure for line-segments, we use a modified Hausdorff distance $d_h(a)$ \cite{Quehl2017SimilarityMeasure}.
In this case, $a$ denotes a match between a detected line-segment and a line-segment landmark.
Here, we only match line-segments which represents the same class of road marking.
For the remaining considerations, we define the generalized feature distance measure
\begin{align}
\label{eq:matching_costs}
d_f(a) = 
\begin{cases}
\phantom{w_{1}} d_e(a) & \text{if pole}\\
w_{h} d_h(a) & \text{if line-segment}\\
\end{cases}
\text{,}
\end{align}
where $w_{h} \in \mathbb{R}$ is a constant factor to weight the line-segment distance relatively to the pole distance.
In the remainder of this section, we describe the RANSAC-based matching which is performed for each $W$.
In each RANSAC iteration, two of all available features in $W$ are randomly selected.
The selected features are randomly associated to two different landmarks in the nearby area of the aerial imagery.
We assume that the selected areas in both domains are partly comprise the same region of the real world based on the rough initial geo-referencing.
Thereafter, we estimate a single transformation $\Delta_G \in SE(2)$ by minimizing 
\begin{equation}
 e_{f}(\Delta_G) = \sum_{i=1}^{N} d_f( a_i )
\label{eq:distance_error}
\end{equation}
using the two selected associations ($N=2$).
For optimization, we use the Levenberg-Marquardt (LM) algorithm \cite{agarwal2010bundle}.
In each LM iteration, the features are transformed with the current estimate of $\Delta_G$ before evaluating (\ref{eq:matching_costs}). 
Afterwards, we transform all features in $W$ with the resulting $\Delta_G$ 
and evaluate (\ref{eq:distance_error}) for all nearest-neighbour matches $a_{nn}$ in $W$ whose distance $d_f(a_{nn})$ is smaller 
as a pre-setted inlier threshold.
This yields the final cost $E_f$ of this iteration. 
The RANSAC loop terminates
until $E_f$ is smaller than a pre-setted threshold $E_{limit}$ or all association combinations are evaluated and no one reached $E_{limit}$.
If $E_{f} < E_{limit}$, the final matches in $W$ are the inliers of the iteration with 
the smallest $E_f$.
Furthermore, we assume that the estimated transformation $\Delta_{G}$ and the transformation $\Delta_{G_{i-1}}$ of the previously estimated 
and partly overlapping window $W_{i-1}$ are similar. 
Therefore, we analyze the transformation difference $\Delta_{G_{i-1}, G} = \Delta_{G_{i-1}}^{-1} \Delta_{G}$ in each RANSAC iteration.
If the angle $\sphericalangle(\Delta_{G_{i-1}, G})$ and the absolute translation $||\mathop{\mathrm{trans}}(\Delta_{G_{i-1}, G})||$ are greater than pre-setted 
thresholds, we reject the solution of the current iteration.
By this, we avoid errorneous transformations caused by e.g. symmetric constellations of feature displacements.
\subsection{Alignment}
\label{subsec:Pose Graph Optimization}
The final step of our pipeline is the joint alignment of all trajectories to the aerial imagery using the final matches from all windows.   
The initialization for this optimization is given through the initially referenced trajectories.
Here, the poses $P$ from all trajectories are the parameters to be adjusted. 
Thereby, each $\mathbf{p} \in P \subset SE(3)$ is a 3D-transformation. 
We minimize
\begin{equation}
 e_a(P) = e_f(P) + w_{\Delta} \sum_{\Delta_{i,j} \in D} \xi \left( \Delta_{i, j}  \mathbf{P}_j^{-1} \mathbf{P}_i \right) 
\label{eq:alignment_error}
\end{equation}
using the LM algorithm. 
The additional sum in (\ref{eq:alignment_error}) penalizes local distortions between nearby poses $\mathbf{P}_i, \mathbf{P}_j \in P$.
Thereby, the $\Delta_{i, j}$ is the original pose difference between $\mathbf{P}_i$ and $\mathbf{P}_j$ from the initialization of $P$ from
which we assume that it already exhibits a high local accuracy and smoothness. 
The topology of the pose difference set $D$ is given from \cite{Sons2018MultiDriveMapping}. 
The $w_{\Delta} \in \mathbb{R}$ is a constant weighting factor and $\xi: SE(3) \rightarrow \mathbb{R}^6$ depicts a 3D-transformation into 
a minimal parameterized representation \cite{sons2015multi}.
Similar to the local optimization in the assocation step, all matched features are transformed based the current estimate of its 
nearest-neighbour pose in $P$ and the given extrinsic sensor calibration in each LM iteration before evaluating $e_f(P)$ in (\ref{eq:alignment_error}).
%
% The whole pose graph optimization can be then formulated as a non-linear optimization problem.
% %
% The error term (cost function), which we want to minimize, is presented through Eq. \ref{eq:Cost-Function}.
% %
% This cost function consist mainly of two part: the first term represent the feature correspondence matching error and the second term represent the delta pose changes error.
% %
% Via minimizing the first error term, we can consider, that the trajectory is accurately geo-referenced to aerial imagery and a global consistency of trajectory is provided.
% %
% Additionally, we add the second error term into the cost function.
% %
% Using the second term, we can minimize the local delta pose changes and avoid much unreliable local pose changes to keep the local consistency of the trajectory unchanged.
% \begin{equation}
%  Cost = \omega_f\sum_{i=1}^{N_{pose}}\sum_{m=1}^{N_f^i} d_f(Fcor_m^i) + \omega_s\sum_{j}^{N_{pose}-1} \Delta P_{j, j+1}
% \label{eq:Cost-Function}
% \end{equation}
% $N_{pose}$ and $N_f^i$ are the number of poses and relevant feature correspondences for pose $i$.
% %
% $w_f$ and $w_s$ are stand for weightings for the first and second error term.
% %
% $\Delta P_{j, j+1}$ presents the delta pose changes from pose $j$ to pose $j+1$.
% %
% $d_f$ is the distance metric for poles and road markings, which we have already used in Sec. \ref{subsec:Feature Association}.
% %
% And $Fcor^i$ are the found feature correspondences relevant to pose $i$.  

%% file: evaluation.tex
\section{Evaluation}
\label{sec:Evaluation}
Within this section, we demonstrate the capabilities of the proposed approach.
In Sec. \ref{subsec:Data Set}, we briefly describe our experimental vehicle and the evaluation dataset comprising three recording sessions.
%
% Firstly, we briefly describe the experiment vehicle and the evaluation data set in Sec. \ref{subsec:Data Set}, which provides three driving passes.
%
Afterwards in Sec. \ref{subsec:Evaluation Method}, we introduce two strategies to evaluate the achieved results.
Finally, we discuss in Sec. \ref{subsec:Discuss of Results} the trajectory alignment for some typical inner-city scenarios.
\subsection{Experimental Setup}
\label{subsec:Data Set}
\begin{figure}
\centering
\includegraphics[width = \columnwidth]{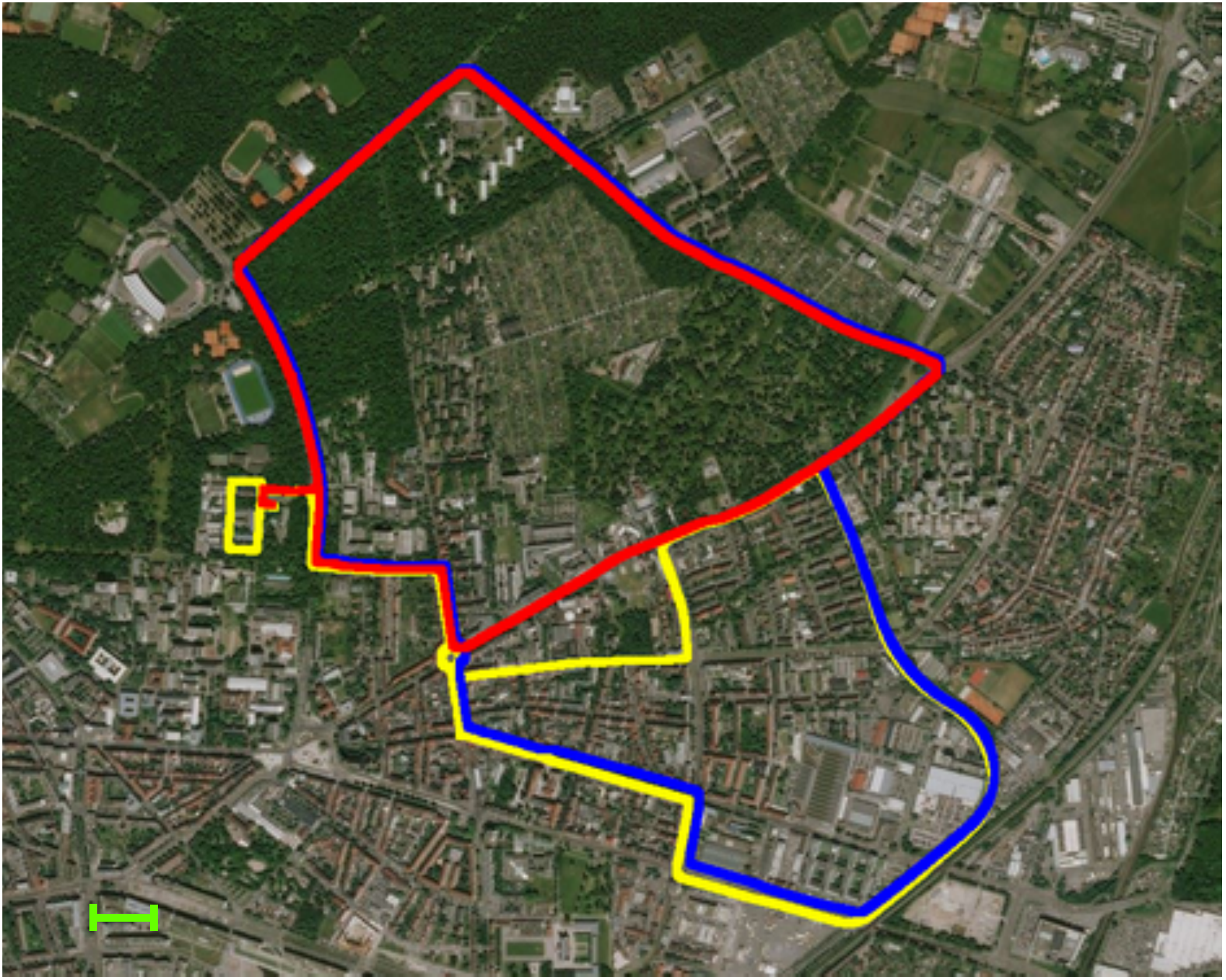}
\caption{
Aerial view of the three jointly aligned and partly overlapping evaluation passes through inner-city and sub-urban area in Karlsruhe. 
The green colored bar corresponds to $100$m.
The entire driven distance is about $19.7$km and each recording session is highlighted with a different color (red: $5.5$km, blue: $7.5$km and yellow $6.7$km).}
\label{fig:passes}
\end{figure}

To evaluate our aerial imagery alignment approach, we recorded data with our experimental vehicle \emph{BerthaOne} \cite{tacs2018making}.
The dataset comprises data recorded from four Velodyne VLP16 LiDARs mounted
flat on the roof, three BlackFly PGE-50S5M cameras behind the front- and rear windshield and a Ublox C94-M8P GNSS receiver.
All sensors are jointly calibrated using the calibration methods proposed in \cite{strauss2014calibrating} and \cite{Kummerle18calib}. 
The recorded data consists of three partly overlapping passes through sub-urban and inner-city streets in the region of Karlsruhe, Germany (see Fig. \ref{fig:passes}).
All sessions are consistently registered and initially referenced with low-cost GNSS measurements using \cite{Sons2018MultiDriveMapping}. 
\subsection{Evaluation Method}
\label{subsec:Evaluation Method}
As for every localization and mapping approach in urban area, 
where also position estimates from post-processed and RTK-corrected high precision GNSS data lacks in accuracy and robustness, 
obtaining reliable ground truth data is a unsolved problem.
%
% Because of the low quality of GPS signal in this area, the post-processed trajectory is mainly not accurate and reliable enough to evaluate our geo-referencing approach.
%
Hence, we evaluate our results with the two following strategies:
\begin{enumerate}
 \item We overlay the initial and the resulting trajectories of our approach to the aerial imagery and analyze carefully every part of the trajectories.
 This gives a visual hint of the alignment accuracy.
 \item To evaluate the results more precisely, we project salient environmental structures from an independet HD-map into the recorded images 
 based on the sensor calibration and the finally estimated trajectories with our approach. 
 The HD-map 
%  bases on the same geo-reference as our used aerial imagery and 
 is provided from an external map provider who generated it by a cumbersome semi-automated mapping procedure.
 By this, we show the alignment accuracy achieved with our approach directly in the recorded images on a pixel-projection level which is highly 
 sensitive to any kind of errorneous alignment.
\end{enumerate}
\begin{figure*} 
\centering
\subfloat[\label{fig:evaluation:a}]{
\includegraphics[width=0.49\linewidth]{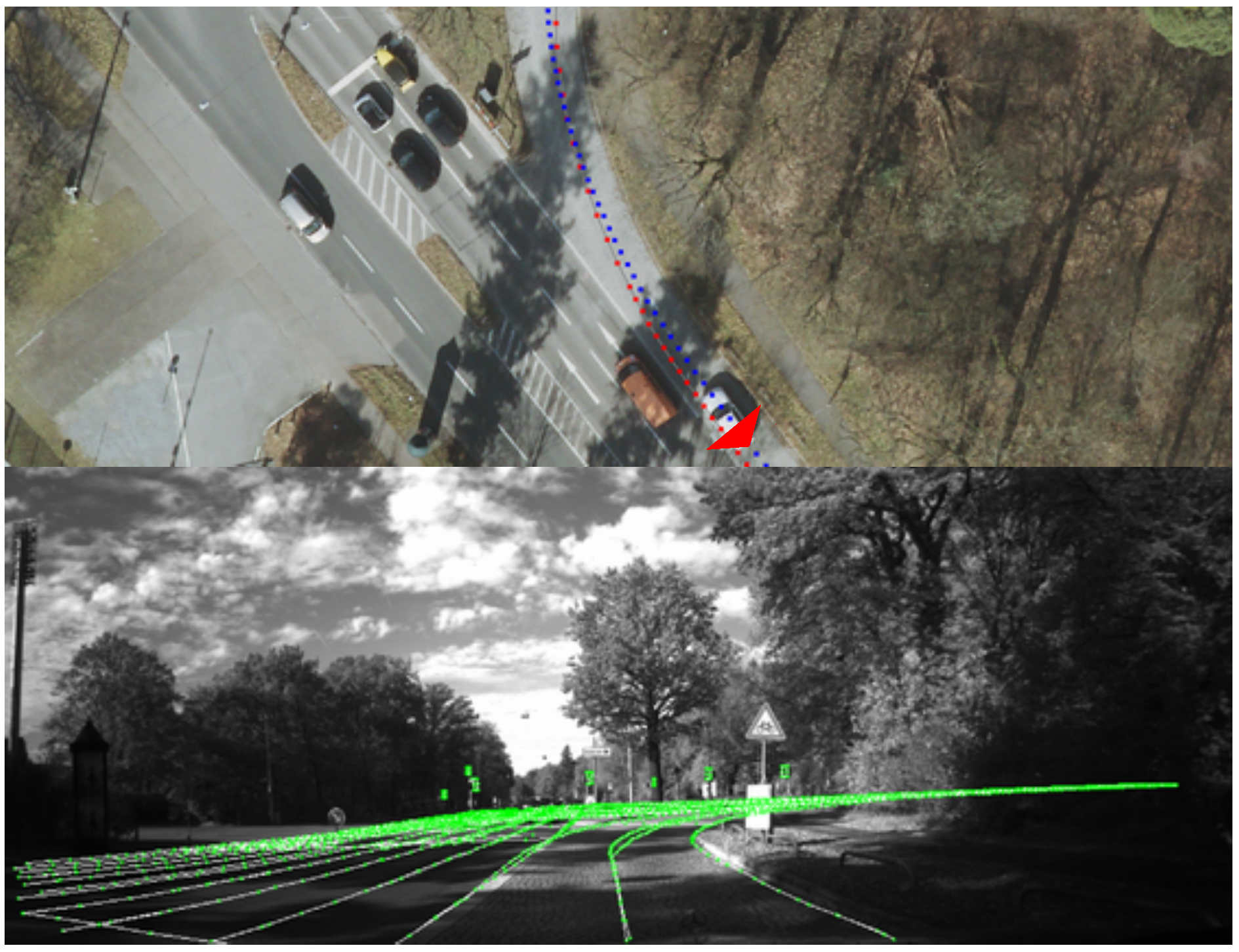}}
\subfloat[\label{fig:evaluation:b}]{
\includegraphics[width=0.49\linewidth]{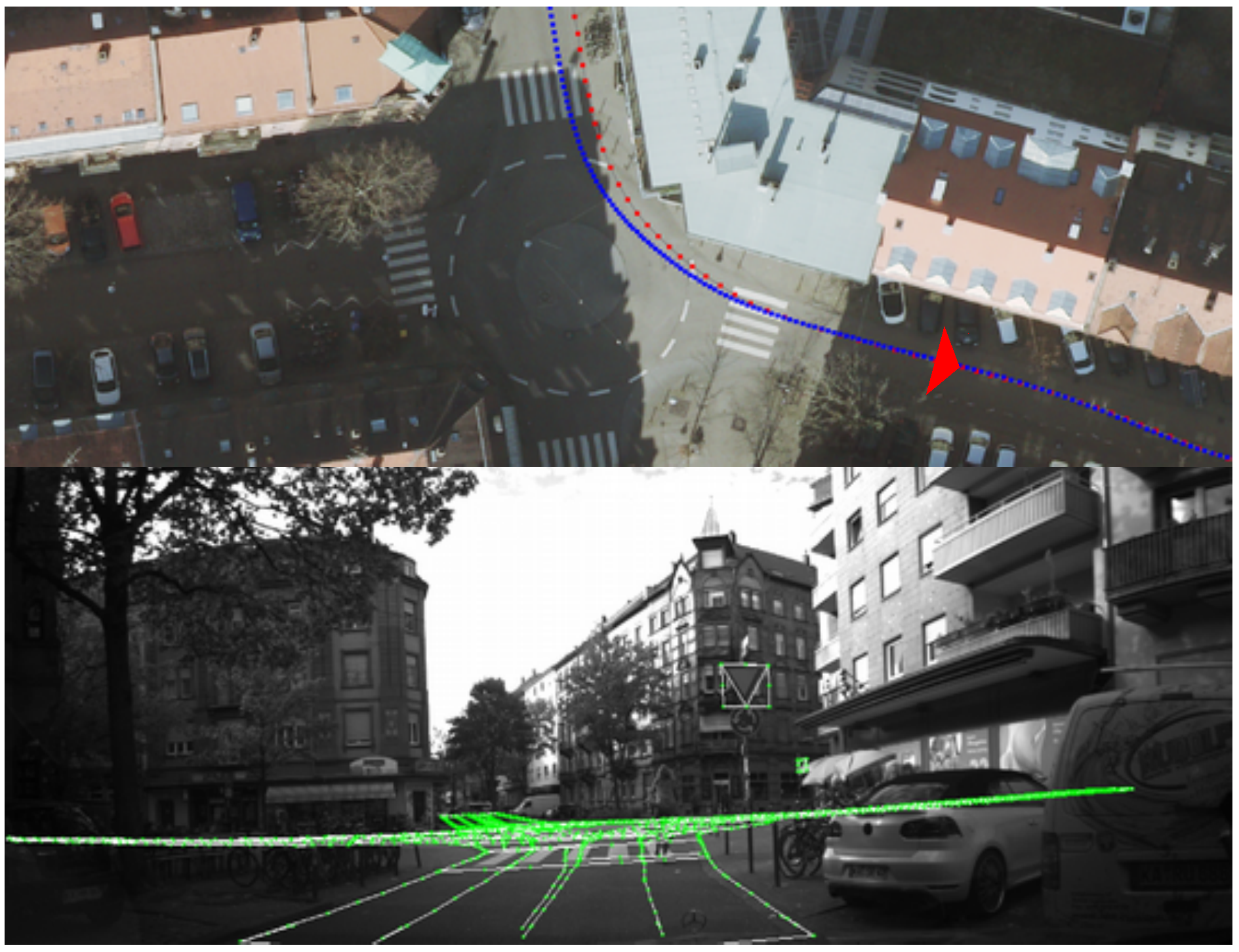}}
\hfill
\subfloat[\label{fig:evaluation:c}]{
\includegraphics[width=0.49\linewidth]{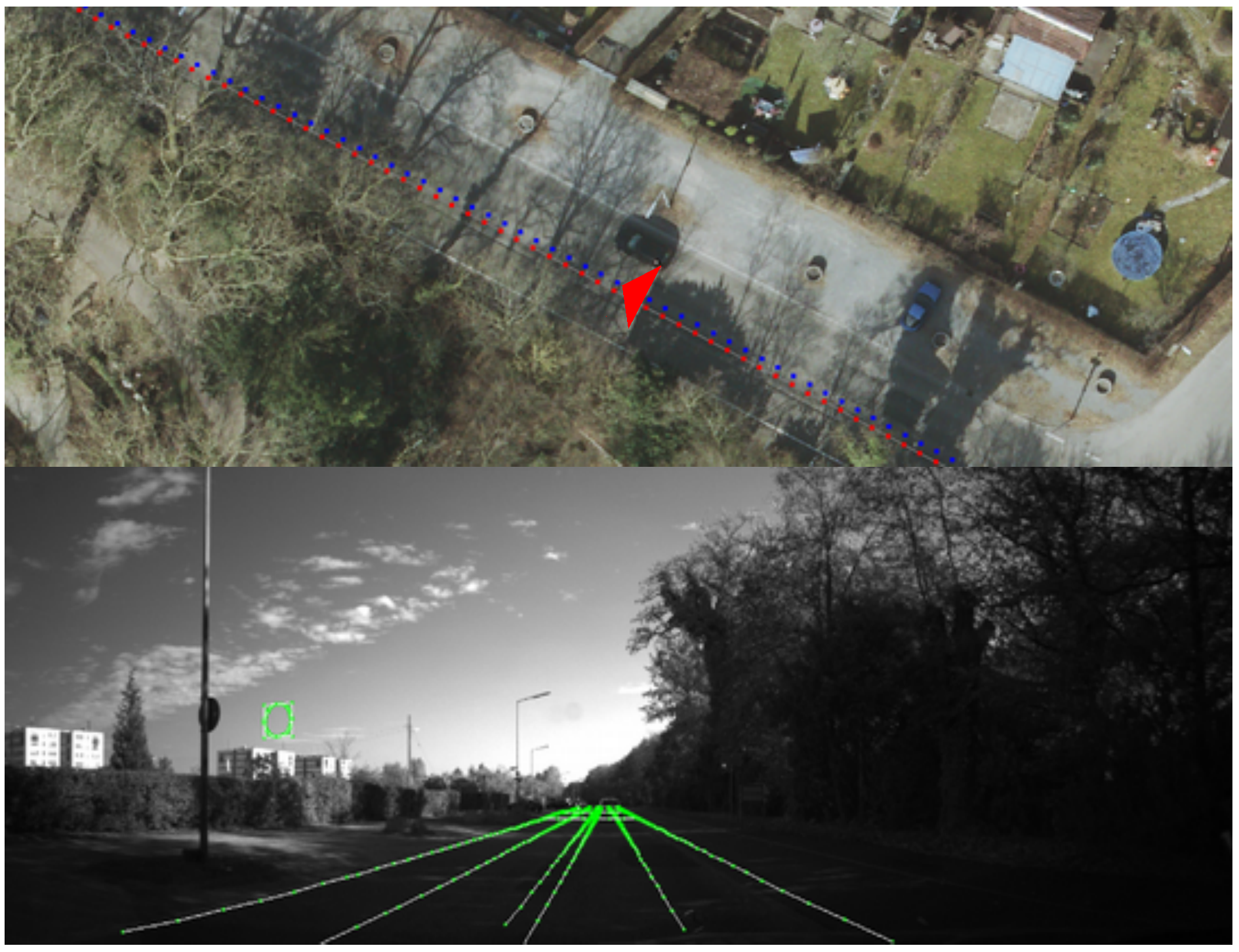}}
\subfloat[\label{fig:evaluation:d}]{
\includegraphics[width=0.49\linewidth]{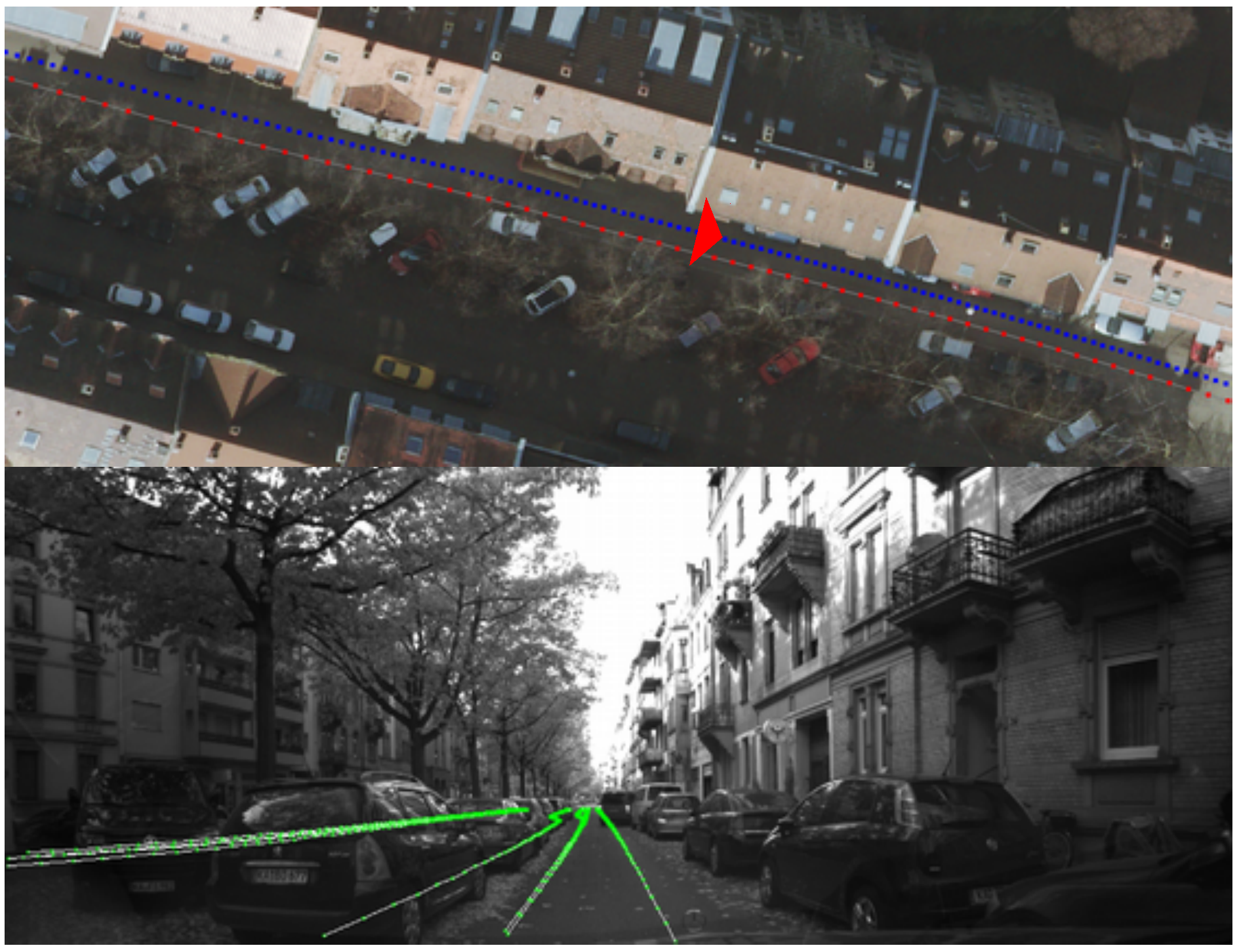}}
\\
\caption{Evaluation results of exemplary scenarios.
The top part of each sub-image shows the initial (red) and the final geo-referenced trajectory (blue) overlayed to the aerial imagery.
The bottom part shows the back projection of mapped structures of the external HD-map into the recorded images.
The red triangle shows the corresponding camera view point.
The top two sub-images \ref{fig:evaluation:a} and \ref{fig:evaluation:b} show good alignment results whereas 
the bottom row sub-images \ref{fig:evaluation:c} and \ref{fig:evaluation:d} show two of the rarely occuring problem cases.
}
\label{fig:evaluation}
\end{figure*}
\subsection{Evaluation Discussion}
\label{subsec:Discuss of Results}
Fig. \ref{fig:evaluation} shows four exemplary scenarios.
The top part of each of the four images shows the initially (red) and our resulting geo-referenced trajectory (blue) overlayed to the aerial imagery.
The lower part shows the back projection of the HD-map into the image of one of the front-facing cameras.
The triangle depicts the view-point of the camera.
The first two sub-images \ref{fig:evaluation:a} and \ref{fig:evaluation:b} show well aligned trajectories. 
From the overlayed trajectories we see that the initially referenced trajectories by GNSS are errorneous since the trajectory is close to the lane boundaries 
which does not explain our driving behaviour while recording the data since we drove on purpose close to the centerline of the lane.  
Furthermore, the global orientation and the scale of the initial trajectory appears errorneous.
In comparison, our resulting trajectory estimate is well aligned to the centerline which coincides with our driving experience. 
Furthermore, the lower image of \ref{fig:evaluation:a} and \ref{fig:evaluation:b} show the reprojection of salient structures (e.g. lane boundaries and -centerlines, traffic lights, road signs)
provided from the independet HD-map into the recorded images using the resulting trajectories. 
Obviously, the projection of the mapped structures show an accurate placement at the correct positions in the images.
Since the projection into images is particular sensitive to angular errors, these reprojections show the high rotation accuracy.
However, also lateral and longitudinal shift errors are clearly visible which occurs rarely if the coverage of features in the area is poor. 
The sub-images \ref{fig:evaluation:c} and \ref{fig:evaluation:d} show exemplary two of those problem cases.
The projection in sub-image \ref{fig:evaluation:c} shows a sufficient lateral and angular alignment which is visible from the well fitting lane boundaries. 
However, a longitudinal error is clearly visible by the projection of the road sign in the left part of the image. 
Here, the poles are not detected in the point clouds due to the high driving velocity and, hence, only the lateral 
error is observable. 
Sub-image \ref{fig:evaluation:d} shows another problem case where road boundaries established by a row of parked cars are recognized.
However, the boundary labels in aerial image fits to the real lane boundaries and, hence, the error minimization leads to a lateral error which is clearly visible. 
Table \ref{tab:Number of extracted poles and road markings} shows the number of detected  and matched features.
%
% Curb lines, dashed lines 12cm and poles have . 
%
The road boundary and dashed line 12cm features have by far the highest detection frequency and are almost equally distrubuted along the passed area 
which enables an overall well lateral alignment (except of the problem case depicted in sub-image \ref{fig:evaluation:d}).
Poles have also a good coverage but occur more frequently at crossings. 
More discriminative road markings like arrows, stop lines or zebra lines occurs mainly close to crossings which leads to 
an excellent alignment at all passed crossings. 
%
% For automated driving applications, crossings are the most challenging scenarios where maps could
%
%
According to our evaluation, $19.1$km of the $19.7$km are properly aligned to the aerial imagery. 
Only at $0.6$km, we recognized errorneous alignment as shown in sub-image \ref{fig:evaluation:c} and \ref{fig:evaluation:d}.
Most of these errors arises from the absence of features and, as a result, a degree of freedom which could not be observed.  
\begin{table}[!htbp]
    \centering
    \begin{tabular}{lccc}
        \hline
        session & red & blue & yellow\\
        \hline
        curb lines & 1282 (89.7\%) & 1627 (92.2\%) & 1854 (94.4\%)\\
        dashed lines 12cm & 408 (80.9\%) & 645 (86.8\%) & 450 (69.3\%)\\ 
        poles & 210 (58.1\%) & 361 (51.8\%) & 163 (61.3\%)\\
        dashed lines 25cm & 46 (82.6\%) & 89 (84.3\%) & 115 (69.6\%)\\
        lines 12cm & 38 (86.8\%) & 36 (83.3\%) & 25 (76.0\%)\\
        arrow lines & 15 (86.7\%) & 19 (78.9\%) & 32 (78.1\%)\\
        lines 25cm & 15 (80.0\%) & 20 (85.0\%) & 14 (78.6\%)\\
        stop lines & 9 (77.8\%) & 11 (72.7\%) & 19 (78.9\%)\\
        pedestrian road lines & 8 (75.0\%) & 9 (77.8\%) & 20 (75.0\%)\\
	zebra lines & 0 (0.0\%) & 14 (78.6\%) & 15 (73.3\%)\\
        bicycle road lines & 3 (100.0\%) & 9 (66.6\%) & 6 (83.3\%)\\
        \hline
    \end{tabular}
    \caption{Number of detected pole and road marking features. 
    The percentage numbers are the share of successfully matched features.}\label{tab:Number of extracted poles and road markings}
\end{table}
%
% Therefore, it's for our approach important, that we can detect poles or separate road markings like stop lines or arrows during driving, with that our approach can get a unique and reasonable feature association.
% %
% In Fig. \ref{fig:evaluation:b}, an latitudinal alignment error is presented, which  is mostly caused by intersection itself.
% %
% In this intersection, there are almost only road boundary features and many cars park beside the lane on the road.
% %
% The boundary of road and no road changes overtime.
% %
% In contrast, we can only label the road boundaries once in aerial imagery.
% %
% Many from our sensor observed features are not the same features, what we have manually labelled in the aerial imagery, which cause the most appeared latitudinal alignment error in our approach.
% %
% Hence, the existence of long-time stable features in intersections is a general requirement to get an accurate trajectory alignment using our approach.

%% file: conclusions.tex
\section{CONCLUSION}
\label{sec:Conclusion_and_future_work}
Within this work, we presented a novel trajectory geo-referencing approach
using salient pole- and road marking features which are well observable in the sensor and the aerial imagery domain. 
We match those features between the two domains robustly by iteratively determining a local displacement error.   
In our real-world experiments, we showed that our approach achieves accurate results in numerous challenging sub-urban and 
inner-city scenarios.
Our experiments proved the importance of both type of features to achieve a sufficient coverage of measurements and 
to observe all degrees of freedom. 
Furthermore, we showed that the alignment using GNSS can not reach this accuracy and robustness due to the lack of reliable 
measurements.
In summary, our approach allows a reliable joint and globally accurate alignment of trajectories from multiple sessions 
in areas where GNSS is not reliable which is a fundamental problem of crowd-map based approaches for automated driving. 

%% file: acknowledgment.tex
\section*{Acknowledgment}
The research leading to these results has received founding from German Federal Ministry of Education and Research (BMBF) as one part of project UNICARagil. The authors would like to thank BMBF for their support.